\documentclass[runningheads]{llncs}

\usepackage[T1]{fontenc}
\usepackage{amsmath}
\usepackage{graphicx}

\begin{document}
%
%
\title{Automatic Extraction of Understandable Controllers from Video Observations of Swarm behaviors}
%
%
\titlerunning{Understandable Controller Extraction from Video Observations of Swarms}

%
%
\author{Khulud Alharthi\inst{1,3}\orcidID{0000-0002-0565-2249}\and
Zahraa S Abdallah*\inst{2}\orcidID{0000-0002-1291-2918} \and
Sabine Hauert*\inst{1,2}\orcidID{0000-0003-0341-7306} }
%
\authorrunning{K. Alharthi et al.}
%
\institute{Bristol Robotics Laboratory, University of Bristol, Bristol, UK \email{khulud.alharthi@bristol.ac.uk} \and Department of Engineering Mathematics, University of Bristol, Bristol, UK
 \and
Department of Computer Science, College of Computers and Information Technology, Taif University, Saudi Arabia\\
*Both authors have contributed equally to the work}

%
\index{Alharthi, Khulud}
\index{Hauert, Sabine}
\index{Abdallah, Zahraa}
%
\maketitle              
\begin{abstract}
Swarm behavior emerges from the local interaction of agents and their environment often encoded as simple rules. Extracting the rules by watching a video of the overall swarm behavior could help us study and control swarm behavior in nature, or artificial swarms that have been designed by external actors. It could also serve as a new source of inspiration for swarm robotics. Yet extracting such rules is challenging as there is often no visible link between the emergent properties of the swarm and their local interactions. To this end, we develop a method to automatically extract understandable swarm controllers from video demonstrations. The method uses evolutionary algorithms driven by a fitness function that compares eight high-level swarm metrics. The method is able to extract many controllers (behavior trees) in a simple collective movement task. We then provide a qualitative analysis of behaviors that resulted in different trees, but similar behaviors. This provides the first steps toward automatic extraction of swarm controllers based on observations.
\end{abstract}
\section{Introduction}
Swarm behavior emerges from simple rules which govern the interaction among the agents and between each agent with their surrounding environment. Birds flocking, fish schooling, and bee foraging are examples of swarm behaviors found in natural systems \cite{10.1007/978-3-540-30552-1_2}. Inspiration has been taken from these natural systems to design robot swarms. Swarm robotics could be used in fire and rescue, storage organization, bridge inspection. Also, in a biomedical application where swarms of large numbers of miniature robots coordinate to detect, monitor, or treat medical conditions \cite{peyer_zhang_nelson_2012,00174,10.3389/frobt.2020.00053}. Swarm behavior is designed by defining the rules of local interaction between agents and their environment \cite{brambilla:hal-01405919}.

Extracting an understandable controller by watching a video of the overall swarm behavior can serve many purposes. It can be considered a design paradigm of swarm robotics by allowing for the automatic extraction of use-able rules for robot swarms just based on a demonstrated behavior of artificial or natural swarms. In addition, the readability of the controller can provide the swarm engineer with the ability to understand, control, or adapt the rules to new robots. Moreover, the method can be used to analyze the natural swarm system. For example, extracting these rules can help behavioral ecology studies interpret how these rules developed and whether the same rules evolved across different species \cite{mann_bayesian_2011}. Another application involves understanding the motion of a particular cell system, which can provide insight into the influence of new medical intervention on this system \cite{ferguson_inference_2016}. This method could also be also used to control natural swarm systems or learn the behavior directly from an online setting \cite{bonnet_robots_2019}. 

Using existing video observations of swarm behaviors as a source of learning has been investigated in several works. In the imitation learning context, a video of an ideal behavior is used to train a controller mostly in form of a neural network to produce a swarm imitating the demonstrated behavior \cite{prorok_holy_2021,8798720,hu2020vgai,8967824,LI2019249,maxeiner_imitation_2019,yu_swarm_2021,10.1007/978-3-642-38715-9_6,erbas_evolution_2015}. For the second context, video observations were used to learn about biological swarm behavior by analyzing their trajectories to investigate what components of the individual controller were crucial for the emergent behavior seen in different species \cite{li_turing_2016,00174,eriksson_determining_2010,mann_bayesian_2011,herbert-read_inferring_2011,amornbunchornvej_framework_2020,schaerf_statistical_2021}.

Most of the proposed works do not provide high-level, understandable rules that can easily be adopted to external systems. In this work, we develop a method to automatically extract understandable controllers from a video observation of swarm behavior. The only input to the proposed method is the video observation of the swarm behavior with no information on the type of the swarm and no requirement of a training dataset. The extracted controller takes the form of a behavior tree to favour human readability \cite{jones2018evolving,hogg_evolving_2020}. An evolutionary algorithm is used to produce a similar emergent behavior to the original observed behavior. The fitness of the evolutionary algorithm is defined as the similarity between two behaviors assessed using eight swarm motion metrics.

The paper is organized as follows. Section 2 introduces an overview of the related works. Section 3 presents the components of the proposed extraction method of the swarm's local controller based on a video observation. Our results are analyzed and discussed in section 4.
\section{Related Works}
The use of a video observation of swarm behavior as a source of learning can be beneficial in several ways. It can be considered an automatic design approach for swarm robotics. Also, it can provide insight into the underlying biological swarm mechanism. Thus, the human readability of the learned controller is highly regarded. In this section, we review works that used video observation to learn a swarm behavior.

Robot swarms are often seen as simple with (a) multiple robots that have (b) simple capabilities and (c) only local perception where (d) all of them collectively work to achieve specific behavior. Designing swarm behaviors that fit these characteristics comes with three important benefits: robustness, ﬂexibility, and scalability \cite{brambilla:hal-01405919,nedjah_review_2019}. Designing the rules that make up a swarm robot's local controller is done either manually or by a careful definition of optimization's objective function to find the rules automatically \cite{brambilla:hal-01405919,10.3389/frobt.2016.00029}. In either case, expertise about the desired swarm behavior is needed. Imitation learning eliminates this requirement and allows the extraction of the swarm controller from a demonstration of the desired emergent swarm performance.

Imitation Learning provides the robots with the ability to learn directly from an expert demonstration. This establishes a learning paradigm between the swarm robots as a learner and the original swarm behavior presented in a video as the teacher \cite{prorok_holy_2021}. Different forms of these three components of imitation learning: the expert swarm demonstration, the learner swarm, and the learning mechanism have been proposed in several works. A teacher could take the form of offline behavioral data consisting of states and actions at different time steps. This behavioral data can be generated from a simulation of the target behavior. Depending on the sensing capabilities of the target behavior, the dataset could include images \cite{8798720,hu2020vgai} or other inferred state descriptors \cite{8967824,LI2019249,maxeiner_imitation_2019}. The teaching data can also be generated by capturing the live behavior from a biological swarm \cite{maxeiner_imitation_2019,yu_swarm_2021}. A robot demonstrator in an online learning environment is another form of the teacher in the imitation learning process \cite{10.1007/978-3-642-38715-9_6,erbas_evolution_2015}. The imitation learning process either aims to train a machine learning model that can map the sensed information into action \cite{8798720,LI2019249}, to translate the observed trajectory into performable form and copy it to the learner robot directly  \cite{8967824,10.1007/978-3-642-38715-9_6}, or to optimize a local controller using inverse reinforcement learning \cite{yu_swarm_2021}. Most of the works conducted in offline imitation learning for swarms focus on extracting swarm controllers in different variations of neural network form,which are known to lack interpretability. These variations include : Graph Neural Network \cite{8967824}, Convolutional Neural Network (CNN) \cite{8798720}, Feed-Forward Neural Network\cite{LI2019249}, Recurrent Neural Network \cite{maxeiner_imitation_2019} or a mixture of them \cite{hu2020vgai}.

Some of the studies that propose methods to extract rules from video observations of swarms have a different goal than imitation learning. While imitation learning aims to gain the extracted rules as a learned task that produce the collective behavior robustly in the same or different environment, these works are only concerned with understanding the mechanism of the behavior observed in the video. Observation-based rule extraction uses parameter fitting techniques such as Bayesian inference, force matching and additive mixture \cite{eriksson_determining_2010,mann_bayesian_2011,herbert-read_inferring_2011}. These techniques receive an input of trajectories from the observed swarm behavior and rules with associated parameters. They then produce values for these parameters to indicate the impact of each of these rules on the emergent behavior. These rules include simple actions such as collision avoidance and aligning velocity. Strategies that the individual of the swarm follow all the time such as following a leader, following its neighbors or only following their own rules have been extracted and tested using fish and baboon observations \cite{amornbunchornvej_framework_2020}\cite{schaerf_statistical_2021}.

Work by \cite{li_turing_2016} proposed a turning-like classifier to differentiate between an imitated sample from the assessed controller and the original sample from the original controller. A co-evolutionary algorithm is used to optimize both the classifier to classify original behavior as original and behavior from the assessed controller as duplicate and also to optimize the assessed controller to deceive the classifier by making it classify their behavior as the original. Another method used a single monitoring robot to infer the parameter of the predefined rule of the swarm in a shared simulation setting \cite{00174}.

Most of the proposed works do not provide high-level, understandable rules that can easily be adopted to external systems. This paper aims to not only imitate the behavior but also to provide human-readability of the local rules that lead to the observed behavior. In addition, no assumptions were made about the demonstrated behavior other than considering it is a swarm behavior. In this work, we aim to extract an understandable swarm controller in the form of a behavior tree with nine motion action nodes from a swarm video observation using an evolutionary algorithm and eight swarm metrics. The extracted behavior tree will explain the swarm observation using the list of leaf nodes built in the system. The no-context aspect of our extraction method with rich options of leaf nodes will make it applicable to a wide range of swarm observations. This work serves as a first step toward extracting more complex behavior trees with different leaf node types such as motion actions, transportation actions, communication, and sensing actions we aim to pursue in future works.
\section{Methodology}
\begin{figure}
\includegraphics[width=\textwidth]{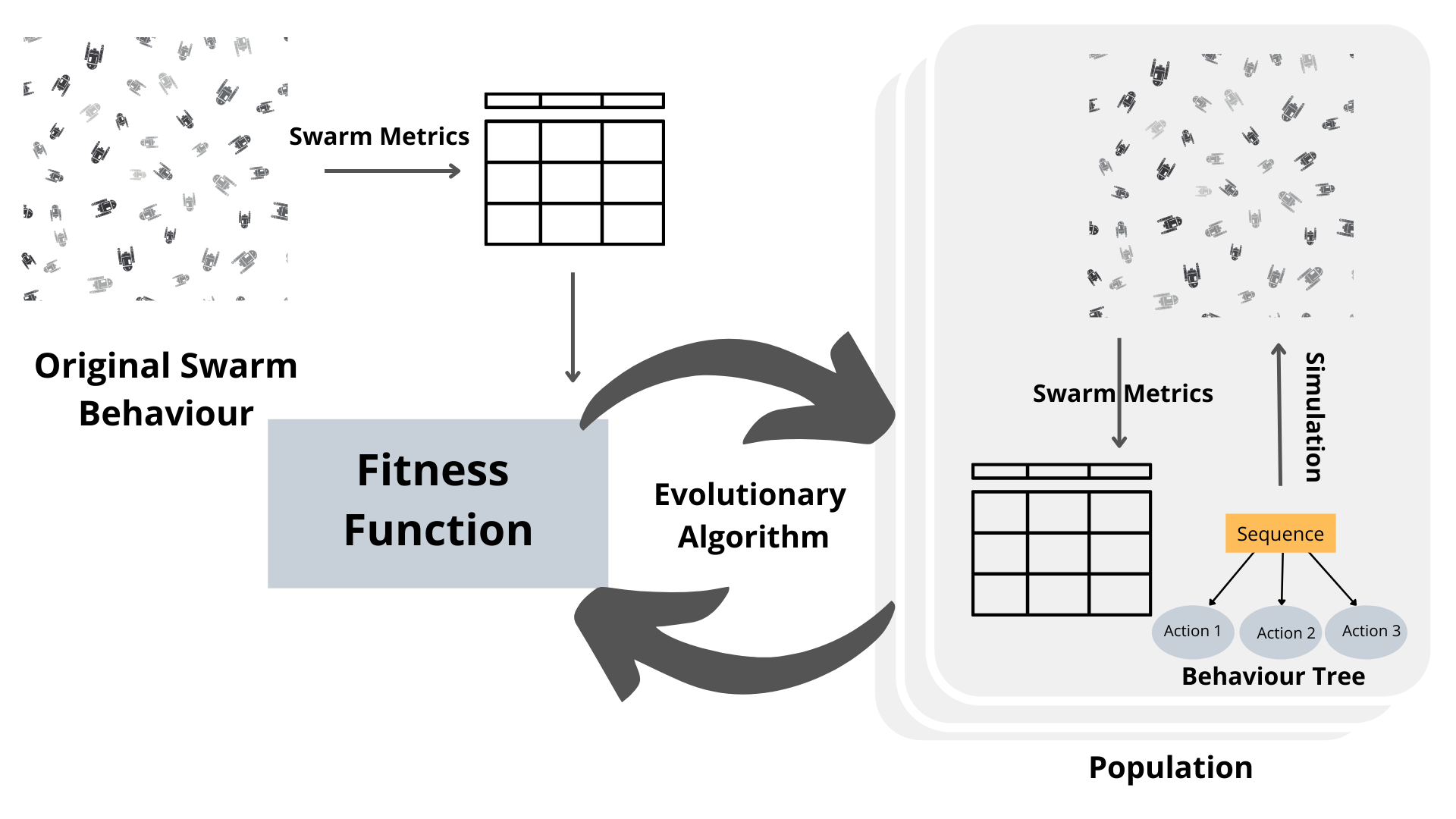}
\caption{Extraction method with artificial evolution to discover behavior trees that generate swarm behaviors with swarm metrics similar to those of the original swarm behavior.} \label{fig1}
\end{figure}

Extracting robot controllers that result in an observed swarm behavior can be defined as an optimization problem with the following formula:
\begin{equation}
\min   distance\left(Original Swarm Metrics,Assessed Swarm Metrics\right) 
\end{equation}
where:
\begin{description}
\item[Original Swarm Metrics] is a vector of metrics describing the swarm motion presented in the video demonstration.
\item[Assessed Swarm Metrics] is a vector of metrics describing the swarm motion generated by simulating a behavior tree to be assessed.
\end{description}
The aim is to minimize the distance between the two vectors, with the assumption that if rules assessed generate swarm behaviors with similar metrics, then imitation has been successful. The controller extraction method starts by computing metrics describing the original swarm behavior. The next steps involve using the evolutionary method to optimize a behavior tree by: for each individual in the initial population, generating random behavior trees, simulating each individual copied over the homogeneous swarm, then measuring the metrics resulting from the swarm behavior, and assigning a fitness based on the distance between this assessed swarm metrics and the original swarm metrics. As generations evolve, high-performing behavior trees are selected, mutated, and crossed over to generate behavior trees that have similar metrics as the original swarm behavior. Figure ~\ref{fig1} shows the general framework of the proposed method.
\subsection{behavior Tree Controller}
Swarm controllers in the form of behavior trees are implemented as a sequence node with three leaf nodes. The sequence node prompts each of its leaf nodes from left to right to execute and return success notifications unless any failure happens in any of the leaf nodes \cite{jones2018evolving}. Leaf nodes can be any of the following nine types:
\begin{itemize}
\item Aggregation: move in the direction of neighboring robots.
\item Dispersion: move away from neighbouring robots.
\item Separation: avoid collision with neighbouring robots.
\item Clustering: move towards the nearest robot to form multiple clusters.
\item Random motion: move in the random direction.
\item South-East force: move to the south-east direction.
\item South-West force: move to the south-west direction.
\item North-East force: move to the north-east direction.
\item North-West force: move to the north-west direction.
\end{itemize}
A library of random controllers is generated to fill the initial population needed by the evolutionary algorithm. Each behavior tree is generated by randomly selecting three of the nine leaf node options.
\subsection{Controller Execution}
To produce the swarm trajectories for each swarm controller, a 2D simulation environment is built using C++ with OpenGL and python Matplotlib. The simulation environment includes a square area (8m x 8m) and 20 swarm agents. Swarm agents are simulated as a circle with a radius of 25 cm where the sensory range of each robot is 50 cm. Agents can move in any direction based on the velocity vector resulting from the execution of the behavior tree with a speed of 1 m/s. Each swarm controller is simulated for 100 time-steps (100 seconds) where at each time step the behavior tree is ticked to update the swarm behavior. 
\subsection{Swarm Metrics}
\label{Swarm Metrics}
The swarm behavior produced from the optimal behavior tree controller should look like the original behavior. Some useful metrics to describe the swarm behavior at a macro level can be found in \cite{Metrics,10.1145/2701973.2702046,hogg_evolving_2020}. Swarm metrics in this work are used to quantify swarm behaviors based on video observations of a swarm. The metrics are computed using only swarm trajectories and are meant to describe a swarm and its resulting trajectories. Four categories of metrics were considered: motion metrics, sparsity metrics, density metrics, and a connectivity metric. A description of these metrics is provided in the following list. In this description, each agent is defined as $a$ where, $a$ is a composite of two vectors that store the location of the agent $a$ in the $x$ and the $y$ direction. $distance$ is computed using Euclidean distance and $n$ is the total number of agents in the swarm. Each of these metrics is a time-series vector as they are computed at each time-step.
\subsubsection{Motion Metrics:}
This category of metrics is used to capture the direction, magnitude and frequency of the swarm motion. It includes three metrics: the center of mass, the maximum swarm shift and the swarm mode index.
\\\textbf{1-Center of mass} is computed as the average overall agent locations in the $x$ and the $y$ direction.
\\\textbf{2-Maximum swarm shift} is computed as the maximum distance moved among all agents measured at each time-step $t$. 
\\\textbf{3-Swarm mode index} is used to measure the frequency of the swarm motion. It is computed as the distance between the center of mass and the swarm mode at each time-step $t$. The swarm mode is defined as a location in the x and the y direction with maximum frequency among all agents' locations. The frequency of location $l$ in the x or the y direction is computed using the following formula  
\begin{equation}
{Frequency(l)}={\underset{ distance(l,l_{i}) < 0.1}{\sum^n_{i=1}1}}
\end{equation}
\subsubsection{Sparsity Metrics:} These metrics describe how sparse the swarm is quantified using two metrics: the longest path and the maximum radius.
\\\textbf{4-Longest path} is the maximum distance traveled from the origin among all agents.
\\\textbf{5-Maximum radius} is defined as the maximum distance among the distances between center of mass of the swarm and each agent. 
\subsubsection{Density Metrics:} Two metrics are used to capture the density, the average local density, and the average nearest neighbour distance.
\\\textbf{6-Average local density} is the sum of the number of agents in the local radius $r$ of each agent averaged over the total number of agents.
\\\textbf{7-Average nearest neighbour distance} is the sum of the distance to the nearest neighbour of each agent averaged over the total number of agents.
\subsubsection{Connectivity Metric:} if the swarm state in each time-step $t$ is considered a graph, with the nodes being the agents, then the connectivity of the swarm can be computed using the \textbf{8-Beta index}. The beta index is a metric that measures the connectivity of the graph by dividing the number of paths between nodes by the number of nodes in the graph. For the swarm beta index, the path is assumed to be connecting two agents if the distance between them exceeds the average distance. Average distance is computed as the sum of the distances among all the agents over the total number of agents.
\subsection{Fitness Function}
The fitness function measures how similar the original swarm behavior is to the assessed swarm behavior based on the swarm metrics extracted from the recorded trajectories. It is defined as the euclidean distance between the original and assessed swarm metric vectors.
\begin{equation}
\sqrt{\sum_{i}(Original Swarm Metrics_{i}-Assessed Swarm Metrics_{i})^{2}}
\end{equation}
Metrics are normalized to ensure each metric contributes equally to the fitness. For each metric, we store the maximum and a minimum values recorded over the whole population of the first generation and use it to normalize over the entire evolutionary run.
\subsection{Evolutionary Algorithm}
Genetic programming (GP) has been used to evolve behavior trees using operations that take into consideration their hierarchical structure \cite{jones2018evolving,hogg_evolving_2020}. In this work, behavior trees are evaluated by the fitness function where the goal of evolution is to minimize fitness. Elitism is used to copy the best three behavior trees to the next generation without any change. The remaining individuals are selected using tournament selection with a tournament size of three. The next steps include applying single-point crossover and single-point mutation with rates as shown in Table\ref{table:1}. In the single-point crossover, the cross point is chosen randomly and the two behavior trees swap their leaf nodes. The Mutation is done by choosing a random leaf node and changing its type to any of the other leaf node types. The behavior tree with the best fitness function in the final generation will then be chosen as the extracted swarm controller.
\begin{table}[h]
\centering
\begin{tabular}{ |p{4cm}|p{3cm}|}
 \hline
 Parameter & Value   \\
 \hline
Population size   &  50  \\
Generations number   & 30  \\
Elitism size   &  3 \\ 
Tournament size  &  3 \\
crossover rate   & 0.5  \\
mutation rate   &  0.3 \\
 \hline
\end{tabular}
\caption{Evolutionary parameters}
\label{table:1}
\end{table}
\section{Results}
In this section, we test the capability of the the swarm metrics to capture the similarity of two swarm trajectories, we then evaluate the performance in correctly extracting original behavior trees. Finally, we provide a qualitative analysis of successful extractions, and extractions resulting in different trees.
\subsection{Evaluation of Swarm Metrics}
\begin{figure}[h]
\makebox[\textwidth]{\includegraphics[width=0.55\paperwidth,height=0.3\textheight]{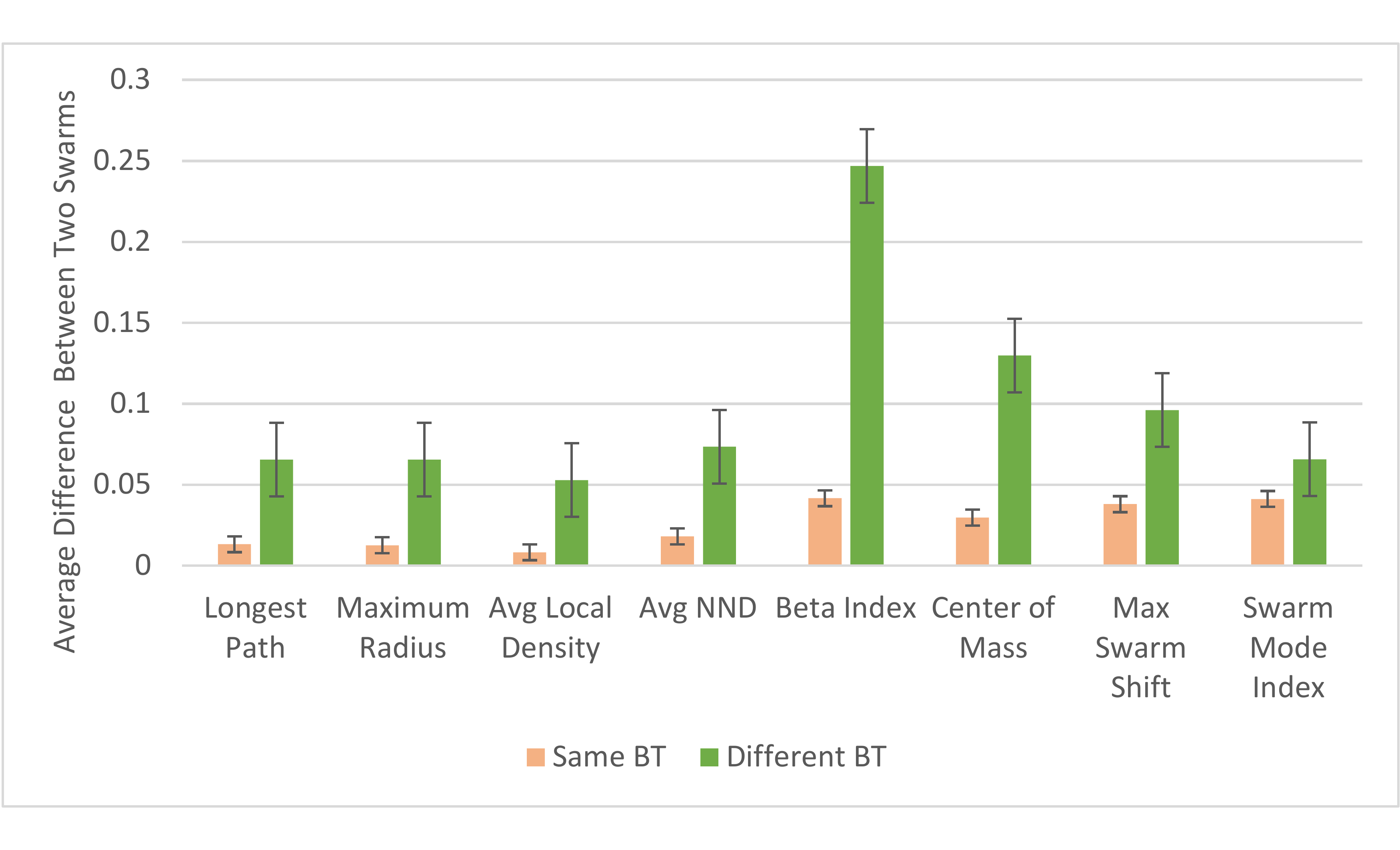}}
\caption{The discrimination power of swarm metrics is apparent when the distance between metrics for swarm behaviors generated from the same behavior trees is smaller than the metrics coming from two different behavior trees.} \label{fig2}
\end{figure}

As a first step, we aimed to see which swarm metrics were useful in differentiating between similar or different behavior trees. To this end, we plot the discrimination power of each metric in section \ref{Swarm Metrics} by comparing their values in two settings. First, the difference between metrics vectors of two swarm trajectories produced by the same controller is computed. In the second setting, the difference was computed based on two swarm trajectories produced by two different randomly generated controllers. 100 pairs of swarm trajectories were used in each setting. Figure ~\ref{fig2} shows the ability of each metric to assess the similarity of two swarm observations. This result shows the distance between metrics for two different swarm behaviors resulting from two different trees is larger than the distance between metrics resulting from the same tree. Thus, demonstrating their discrimination potential. Although some metrics show a higher capability to discriminate than others, each of them can provide a different contribution to the fitness function. 
\subsection{Performance of The Controller Extraction Method }
\begin{figure}[h]
\begin{center}
   \makebox[\textwidth]{\includegraphics[width=0.53\paperwidth,height=0.35\textheight]{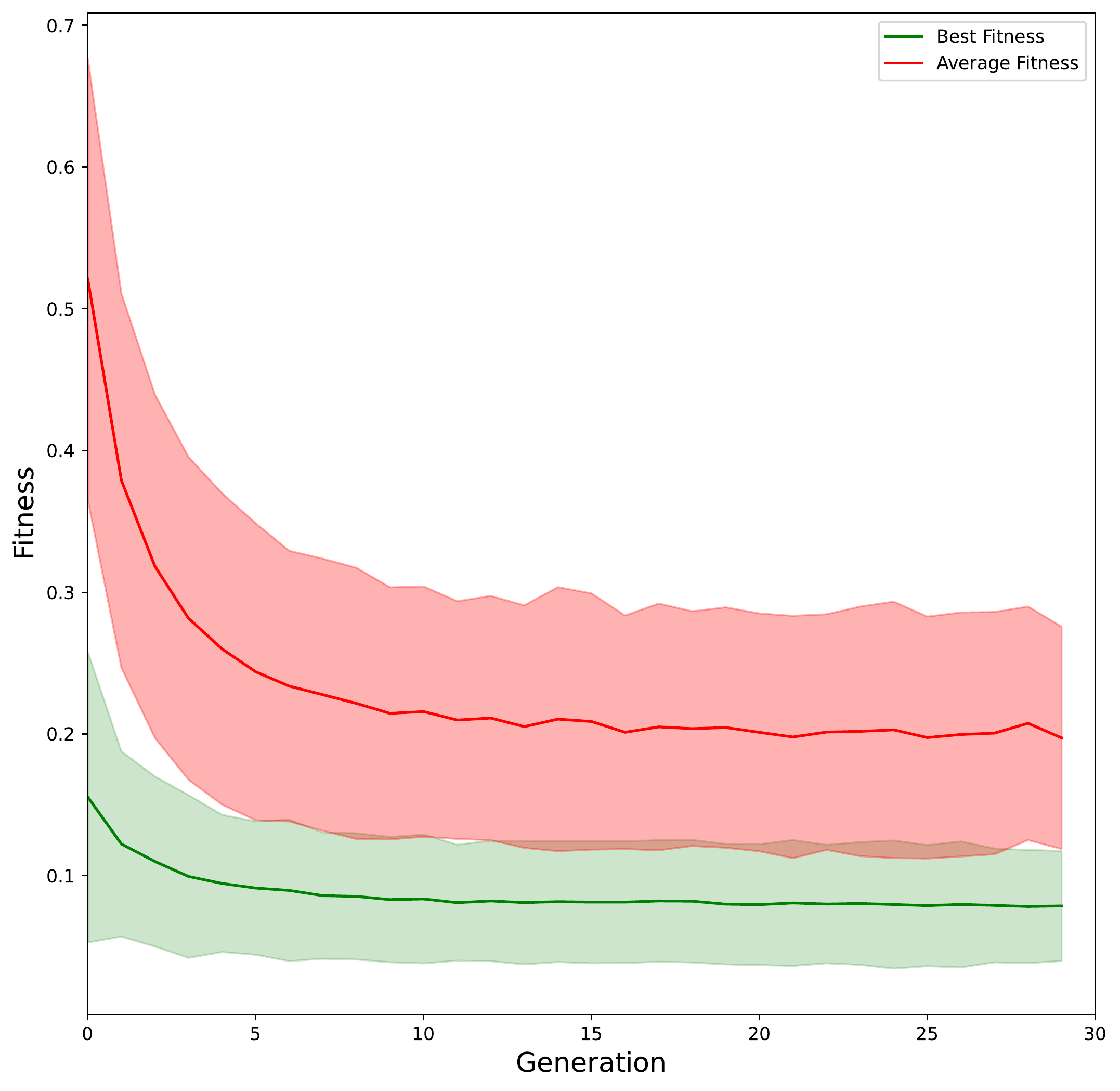}}
\end{center}

\caption{The learning curve of the best fitness (in green) and the average fitness (in red) over the 30 generations shows a successful minimization of the distance between the original and the assessed swarm behavior metrics.} \label{fig3}
\end{figure}

To quantify the performance of the extraction method, we randomly generate 100 behavior trees, each one used as the original swarm behavior from which a behavior tree needs to be extracted. To have a meaningful behavior, generations of the original behavior trees were constrained by preventing leaf nodes that have a canceling effect on each other from being presented in the same behavior tree such as aggregation node and dispersion node. The simulated swarm trajectories of these behavior trees are then used as an input to the controller extraction method.
A Jaccard index is used to evaluate the produced controller by assessing the similarity of the extracted controller to the original behavior tree. This metric is used as ground truth to assess the exact similarity between behavior trees, and can not be used in the fitness which captures indirect metrics that can be extracted from the video observations. The Jaccard index is a measure of similarity between two texts and is computed by dividing the intersection of characters between the two strings over the union of all the characters \cite{JD}. When the Jaccard index is one, the two strings are the same, whereas zero indicates they are completely different. Here behavior trees are represented as strings. The method was able to achieve 0.868 average Jaccard similarity over the 100 behavior trees where 75 behavior trees out of 100 were the exact copy of the original behavior trees. The rest can be grouped into two groups. A high similarity group with a Jaccard index larger than or equal to 0.5 which includes 18 extracted controllers. The last 7 controllers have a Jaccard index less than 0.5 and are in the low similarity group. However, no controller was extracted with a zero similarity. That means the extracted controller in the worst-case will have at least one of the nodes the same as the original controller. Table \ref{tab2} shows a summary of the method's performance.
\begin{table}[h]
\centering
\caption{Performance measures of the controller extraction method}\label{tab2}
\begin{tabular}{c c}
\hline
Accuracy &  75\\
\hline
controllers with high similarity  & 18\\
\hline
controllers with low similarity  & 7\\
\hline
Average Jaccard Index(all)  &  0.868 \\
\hline
\end{tabular}
\end{table}
There is a significant improvement over the first generations as shown by the best fitness of all the 100 behavior trees in Figure ~\ref{fig3}. The average fitness also demonstrates some learning but with larger distribution values than the best fitness.
Increasing the number of leaf nodes of the behavior tree will increase the search space as well as the complexity of the problem. The performance of this method was tested against more complex behaviors coming from behavior trees with 4, 5 and 6 leaf nodes. The results obtained show the potential of this method as it was able to extract 70 exact behavior trees with four leaf nodes, 57 behavior trees with five and 43 with six leaf nodes. Although with 6 leaf nodes, the Jaccard similarity was 0.778, no controller was extracted with zero similarity. For each of these 43 extracted controllers, the method was able to search around 3,000 possibilities of behavior trees, which is not trivial space.
\subsection{Qualitative behavioral Analysis}
\begin{figure}[h]
  \centering
  \begin{tabular}{cc}
  \begin{tabular}{cc}
        \includegraphics[width=0.13\paperwidth,height=0.13\textheight]{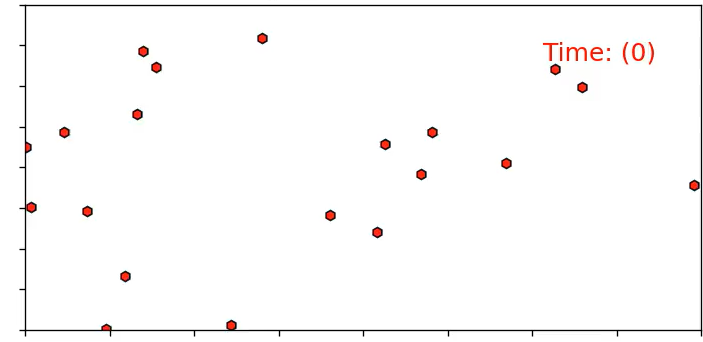} &
        \includegraphics[width=0.13\paperwidth,height=0.13\textheight]{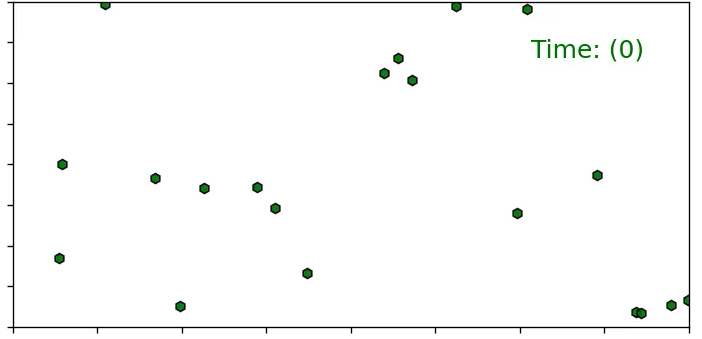}
        
\end{tabular}
   
    &
     \begin{tabular}{cc}
        \includegraphics[width=0.13\paperwidth,height=0.13\textheight]{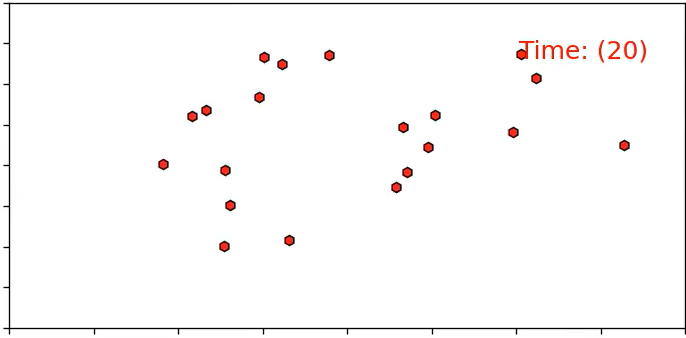} &
        \includegraphics[width=0.13\paperwidth,height=0.13\textheight]{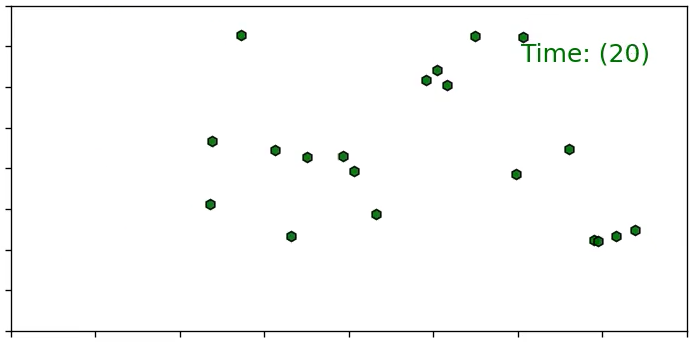}
        
\end{tabular}
    \\
 \small (Initial setting) &     \small (After 20 seconds)
 \\
  \begin{tabular}{cc}
        \includegraphics[width=0.13\paperwidth,height=0.13\textheight]{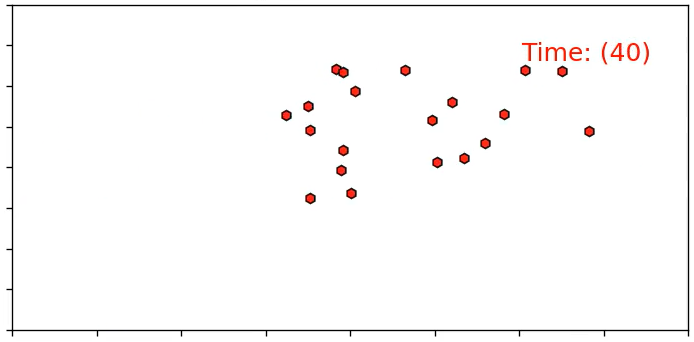} &
        \includegraphics[width=0.13\paperwidth,height=0.13\textheight]{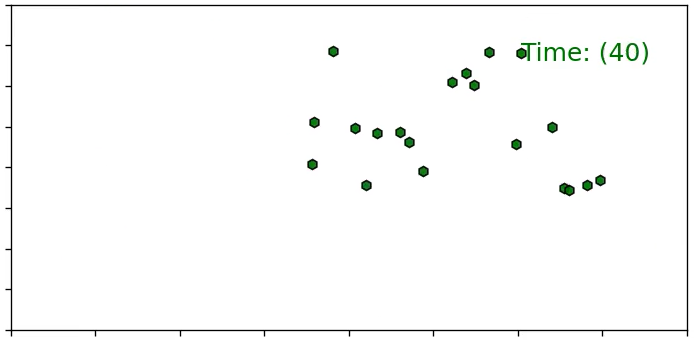}
        
\end{tabular}
   
    &
      \begin{tabular}{c}
        \includegraphics[width=0.26\paperwidth,height=0.13\textheight]{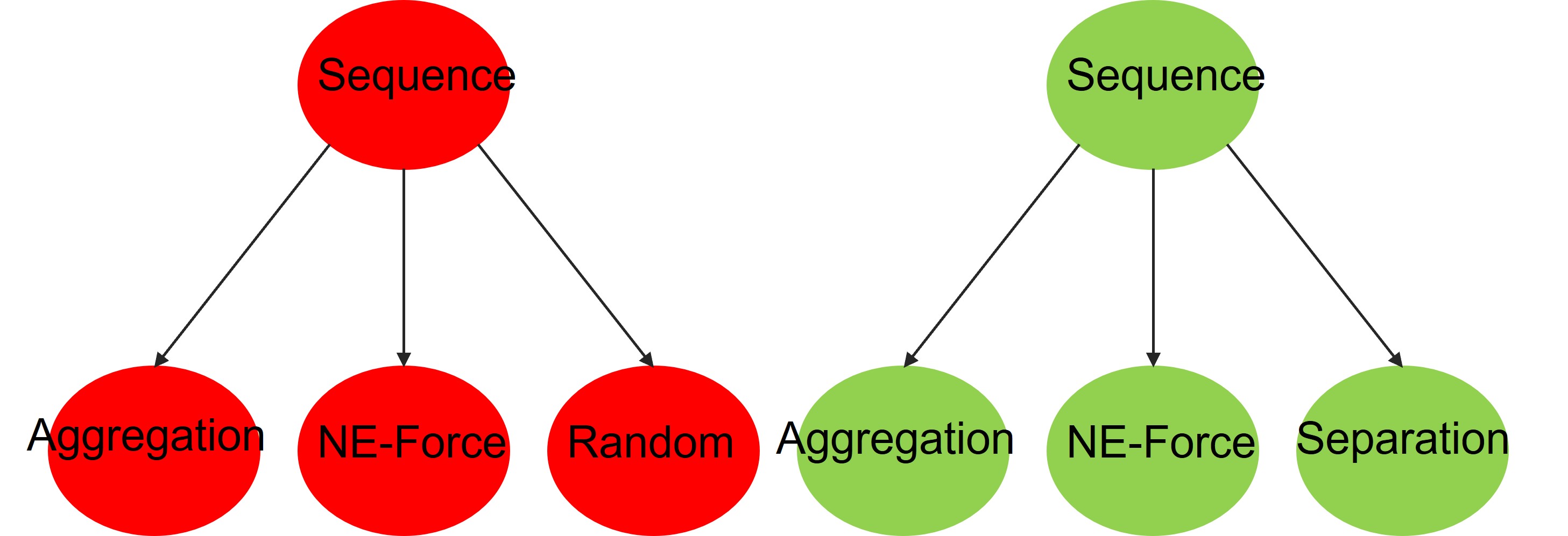} 
        
\end{tabular}
  
    \\
 \small (After 40 seconds) &     \small (Original and extracted behavior trees)
    
  \end{tabular} \qquad 
  \caption{An example of the output of the method with a high Jaccard similarity includes three screenshot of the controller and the behavior from the right to the left. The original behavior tree and their simulated behavior are shown in red whereas the extracted behavior tree and the imitated behavior are presented in green.}
   \label{fig4}
\end{figure}

An example of a produced controller along with the original controller is presented in Figure \ref{fig4}. The original behavior tree includes : random motion node, aggregation node and North-East node. Aggregation node and North-East node were extracted successfully while random node was not. Overall, the random node faced a failed extraction 22 times out of 37. In 18 cases, the random node was confused with a separation node as shown in this example. This is not surprising since combining separation node and leaf nodes with opposite behavior such as aggregation and clustering could look similar to the motion of the random node. Separation nodes itself were wrongly extracted in just three cases out of 26. Aggregation, dispersion, clustering and all the directional nodes except north-east node had zero failed extractions. In general, the directional nodes tend to be over-extracted even when they are not present in the original controller.

\section{Conclusion}
Swarm behaviors in natural systems are inspiring in terms of their ability to provide these systems with robustness and flexibility. Extracting the rules from such systems is crucial for both the engineering of swarm robotics and the interpretability of the underlying dynamics of the swarm systems whether natural or artificial. Such extractions are also useful to understand and control artificial systems after observation. In this study, we developed an understandable swarm controller extraction method using an evolutionary algorithm and eight swarm metrics. To evaluate the method, we constructed a swarm dataset where each sample contains a behavior tree as swarm controller and corresponding simulated swarm trajectories. Our experimental results show the method can exactly extract 75 behavior trees out of 100 behavior trees while obtaining a 0.868 averaged Jaccard similarity. These results show the potential of the method in applications ranging from robotics to biology. In the future, more complex behavior trees and more action nodes will be considered such as transportation, communication and sensing nodes.

%
%
%
%
\bibliographystyle{splncs04}
\bibliography{mybibliography}
%
%




%
\end{document}